\documentclass[letterpaper,10pt,conference]{IEEEtran}

\usepackage{cite}
\usepackage{amsmath,amssymb}
\usepackage{algorithm}
\usepackage{algorithmic}
\usepackage{graphicx}
\usepackage{booktabs}
\usepackage{multirow}
\usepackage{url}
\usepackage{hyperref}
\usepackage{xcolor}
\usepackage{adjustbox}
\usepackage[letterpaper, top=0.75in, bottom=0.75in, left=0.75in, right=0.75in]{geometry}
\usepackage{framed}

\makeatletter
\let\old@maketitle\@maketitle
\def\@maketitle{\vspace*{0.25in}\old@maketitle}
\makeatother
\def\BibTeX{{\rm B\kern-.05em{\sc i\kern-.025em b}\kern-.08em
    T\kern-.1667em\lower.7ex\hbox{E}\kern-.125emX}}

\newif\ifanon
\anontrue

\title{NGPS: GPS-Denied Aerial Geo-Localization and 2.5D Reconstruction via Deep Satellite Image Matching and Multi-Rate Sensor Fusion}

\author{\IEEEauthorblockN{Sanket Sharma}
\IEEEauthorblockA{\textit{Independent Researcher} \\
Email: sharma.sanket272@gmail.com}
}

\begin{document}

\onecolumn
\thispagestyle{empty}

{\LARGE \textbf{IEEE copyright notice}}

\vspace{0.4in}
\noindent
© 2026 IEEE. Personal use of this material is permitted. Permission from IEEE must be obtained for all
other uses, in any current or future media, including reprinting/republishing this material for advertising
or promotional purposes, creating new collective works, for resale or redistribution to servers or lists, or
reuse of any copyrighted component of this work in other works.

\vspace{0.3in}
\noindent
Accepted to be published in \textit{2026 IEEE/RSJ International Conference on Intelligent Robots and
Systems (IROS)}, Pittsburgh, PA, USA, September 27 -- October 1, 2026

\vspace{0.3in}
\noindent
Cite as:

\vspace{0.1in}
\noindent
\fbox{\begin{minipage}{0.95\textwidth}
\vspace{0.05in}
S. Sharma, ``NGPS: GPS-Denied Aerial Geo-Localization and 2.5D Reconstruction via Deep Satellite
Image Matching and Multi-Rate Sensor Fusion,'' in \textit{2026 IEEE/RSJ International Conference on
Intelligent Robots and Systems (IROS)}, Pittsburgh, PA, USA, September 27 -- October 1, 2026, to be
published.
\vspace{0.05in}
\end{minipage}}

\vspace{0.5in}
\noindent
\textbf{BibTeX:}

\vspace{0.1in}
\noindent
\begin{minipage}{0.95\textwidth}
\begin{framed}
\begin{verbatim}
@inproceedings{sharma_ngps_2026,
   author = {Sharma, Sanket},
   booktitle = {2026 IEEE/RSJ International Conference on Intelligent Robots
                and Systems (IROS)},
   title = {NGPS: GPS-Denied Aerial Geo-Localization and 2.5D Reconstruction
            via Deep Satellite Image Matching and Multi-Rate Sensor Fusion},
   address = {Pittsburgh, PA, USA},
   year = {2026},
   publisher = {IEEE, to be published}
}
\end{verbatim}
\end{framed}
\end{minipage}

\newpage
\twocolumn

\maketitle

\begin{abstract}
\small

We present NGPS (Next-Generation Positioning System), a visual geo-localization framework for high-altitude UAVs that provides GPS-free absolute positioning by matching down-facing images to georeferenced satellite imagery with deep features. The system combines (1) adaptive confidence-weighted UKF fusion, where NGPS covariance is modulated by RANSAC inlier ratio, reprojection error, and match confidence; (2) velocity-predictive kernel extraction, using VIO velocity to predict the satellite search region; and (3) an asynchronous multi-rate temporal priority queue that interleaves absolute position (1--2\,Hz), VIO (10--20\,Hz), and IMU (100--200\,Hz) updates in chronological order. Globally optimized poses from VINS pose-graph optimization, anchored by NGPS corrections, further enable real-time 2.5D georeferenced orthomosaic reconstruction. On five flight sequences (60--150\,m AGL), NGPS achieves 2.94\,m position RMSE, with worst-case ATE 6.04\,m at 150\,m AGL and 2\,m/s, yielding a $3.5\times$ improvement over standalone monocular VIO. The system runs in real time on an NVIDIA Jetson Orin NX. Part of the implementation is open-sourced at \url{https://github.com/snktshrma/ngps_flight}.

\normalsize
\end{abstract}


\section{Introduction}

Accurate geo-localization is fundamental to autonomous UAV operations, enabling precise navigation, path planning, and mission execution. While Global Navigation Satellite Systems (GNSS) provide reliable positioning in open environments, performance degrades significantly in urban canyons, under intentional jamming, or in contested environments. For high-altitude drone applications - including surveillance, mapping, search and rescue, and infrastructure inspection - robust GPS-free positioning is increasingly critical as UAVs are deployed in environments where GPS reliability cannot be guaranteed.

Visual-Inertial Odometry (VIO) systems \cite{vins_mono, orbslam3} have emerged as the primary solution for GPS-denied navigation, providing high-frequency relative pose estimates by fusing camera and IMU measurements. However, VIO suffers from inevitable drift accumulation: errors in relative motion estimates compound without external corrections, and loop closure mechanisms are ineffective in exploratory missions or large-scale operations where revisiting locations is infrequent.

This paper presents NGPS, a visual geo-localization framework for high-altitude drones (60--150+\,m above ground level, AGL) with down-facing cameras. At sufficient altitude the aerial perspective resembles satellite imagery, enabling deep feature matching for absolute position without incremental mapping or loop closures; globally optimized poses enable real-time 2.5D orthomosaic reconstruction~\cite{nex_uav}.
NGPS addresses several challenges unique to aerial-to-satellite matching at high altitudes (60--150\,m): (1)~scale and resolution differences between satellite and camera imagery, (2)~viewpoint variations from drone attitude changes, (3)~temporal scene changes in reference imagery, and (4)~computational constraints on embedded platforms.

The main contributions of this work are:
\begingroup
\setlength{\itemsep}{0.2em}
\setlength{\topsep}{0.2em}
\setlength{\partopsep}{0pt}
\setlength{\parskip}{0pt}
\begin{enumerate}
    \item An adaptive confidence-weighted sensor fusion architecture where NGPS measurement noise is dynamically modulated by matching quality metrics (RANSAC inlier ratio, reprojection error, LightGlue confidence), enabling the UKF to aggressively correct drift from high-quality matches while discounting unreliable ones, improving RMSE by 11\% over fixed-noise fusion.
    \item A velocity-predictive kernel extraction strategy that uses UKF-estimated velocity to forward-predict the satellite search region between NGPS updates, reducing matching failures by 32\% at 4\,m/s flight speed.
    \item An asynchronous multi-rate temporal priority queue that optimally interleaves absolute position (1--2\,Hz), VIO (10--20\,Hz), and IMU (100--200\,Hz) measurements in strict chronological order for UKF fusion.
    \item A 2.5D georeferenced orthomosaic reconstruction using globally optimized poses from VINS pose graph optimization anchored by NGPS corrections, enabling real-time terrain mapping.
    \item Real-time implementation on embedded hardware (Jetson Orin NX) demonstrating 2.94\,m geo-localization RMSE across five sequences (60--150\,m AGL), with a $3.5\times$ improvement over monocular VIO.
\end{enumerate}
\endgroup

\section{Related Work}

\subsection{Visual Localization for Aerial Vehicles}

VIO systems address GPS-denied navigation by fusing camera and IMU measurements. Early monocular visual odometry approaches suffer from scale ambiguity and drift. VINS-Mono~\cite{vins_mono} introduced tightly coupled optimization-based monocular VIO with IMU preintegration, employing a sliding window framework that maintains recent camera poses while marginalizing older states for real-time performance. MSCKF~\cite{msckf} provides efficient EKF-based visual-inertial fusion. ORB-SLAM3~\cite{orbslam3} supports visual-inertial multimap SLAM with loop closure. OKVIS~\cite{okvis} performs keyframe-based visual-inertial SLAM using nonlinear optimization. These systems provide accurate short-term estimates but require loop closures or external corrections to bound long-term drift. VINS-Fusion~\cite{vins_fusion} extends VINS-Mono to multi-sensor fusion including GPS, but assumes GPS availability for global corrections.

\subsection{Deep Feature Matching}

Traditional handcrafted feature descriptors such as SIFT, SURF, and ORB struggle under significant appearance changes, scale variations, and illumination differences. SuperPoint~\cite{superpoint} uses self-supervised training to learn both keypoint detection and description, achieving strong performance across viewpoint and illumination changes. Building on learned matching approaches like SuperGlue, LightGlue~\cite{lightglue} employs adaptive inference depth that stops early on easy image pairs, point pruning that discards unmatchable points at early layers, rotary positional encoding for relative position representation, and bidirectional cross-attention for efficient correspondence estimation, achieving real-time performance on embedded hardware.

\subsection{Aerial and Satellite Image Matching}

Cross-view geo-localization, particularly between ground-level and satellite imagery, has received significant attention~\cite{lin_satellite, shi_crossview, huang_crossview}. However, aerial-to-satellite matching presents fundamentally different challenges, as both views share an approximate top-down perspective but differ in resolution, temporal appearance, and viewing angle. Prior work has explored feature-based~\cite{kim_satellite} and learning-based~\cite{chen_satellite} approaches for aerial image registration. UAV-based photogrammetry has also been extensively studied for 3D mapping applications~\cite{nex_uav}. Our work differs by focusing specifically on real-time high-altitude drone localization and integrating matching results into a complete navigation and 2.5D reconstruction pipeline with multi-rate sensor fusion.

\subsection{Sensor Fusion for State Estimation}

Sensor fusion for robot state estimation commonly employs Extended Kalman Filters (EKF) or Unscented Kalman Filters (UKF)~\cite{ukf}. The EKF linearizes the system model around the current estimate, which can introduce errors for highly nonlinear systems, while the UKF uses sigma point sampling to capture nonlinear effects more accurately. Existing multi-sensor fusion frameworks typically assume synchronized measurements or handle multi-rate fusion through simple interpolation or decimation. Our architecture specifically addresses the challenge of integrating low-frequency absolute position measurements (1--2\,Hz) from visual geo-localization with high-frequency VIO data (10--20\,Hz) through an asynchronous temporal priority queue mechanism, ensuring optimal state estimation despite significant rate disparities. The UKF output is further integrated with ArduPilot's high-rate EKF (100--400\,Hz) for final flight control state estimation.

\section{Methodology}

\subsection{System Overview}

The NGPS system comprises three modules within a ROS2 framework, integrated with ArduPilot:

\begin{enumerate}
    \item \textbf{NGPS Module:} Visual geo-localization via deep feature matching against satellite imagery, providing absolute position at 1--2\,Hz.
    \item \textbf{VIO Module:} Tightly coupled visual-inertial odometry with IMU preintegration and sliding window optimization~\cite{vins_mono}, providing relative pose at 10--20\,Hz.
    \item \textbf{UKF Fusion Module:} Asynchronous multi-rate fusion producing drift-corrected odometry at 10--20\,Hz, which feeds ArduPilot's EKF operating at 100--400\,Hz for final state estimation.
\end{enumerate}

\begin{figure}[t]
    \centering
    \includegraphics[width=\columnwidth]{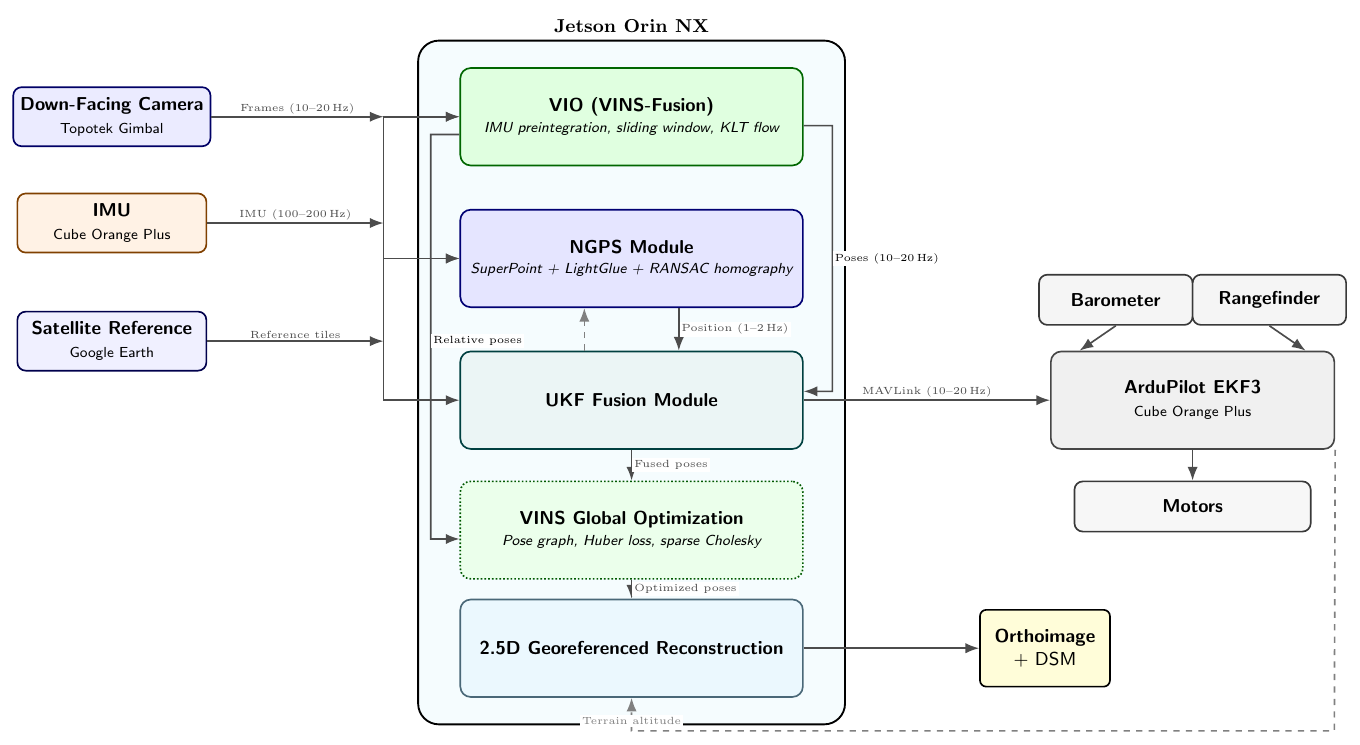}
    \caption{NGPS system architecture: visual geo-localization provides absolute corrections (1--2\,Hz) fused with VIO (10--20\,Hz) via UKF, feeding ArduPilot's EKF (100--400\,Hz).}
    \label{fig:system_architecture}
\end{figure}

This two-stage architecture allows computationally expensive visual geo-localization (1--2\,Hz) to provide periodic absolute corrections while maintaining high-frequency pose estimates for responsive flight control. Figure~\ref{fig:system_architecture} illustrates the data flow. Table~\ref{tab:parameters} summarizes the key system parameters.

\begin{table}[t]
\centering
\caption{Key System Configuration Parameters}
\label{tab:parameters}
\footnotesize
\begin{tabular}{lcp{3.2cm}}
\toprule
\textbf{Parameter} & \textbf{Value} & \textbf{Description} \\
\midrule
\texttt{kernel\_size} & 300 & Reference kernel size (px) \\
\texttt{match\_threshold} & 0.5 & LightGlue confidence threshold \\
\texttt{min\_matches} & 20 & Min. valid correspondences \\
\texttt{max\_keypoints} & 2048 & Max keypoints per image \\
\texttt{max\_rot\_change} & $30^\circ$ & Frame-to-frame rot. limit \\
\texttt{rot\_std\_thresh} & $15^\circ$ & Rotation variance limit \\
\texttt{rot\_history} & 10 & Smoothing buffer size \\
\texttt{ukf\_rate} & 10\,Hz & UKF output frequency \\
\bottomrule
\end{tabular}
\end{table}

\subsection{Visual-Inertial Odometry Module}

The VIO module is based on VINS-Fusion~\cite{vins_fusion}, providing tightly coupled monocular-inertial odometry optimized for high-altitude flight. IMU measurements between consecutive camera frames are preintegrated in the body frame following~\cite{vins_mono}, enabling efficient sliding window bundle adjustment over a window of recent camera poses (typically 10 keyframes) with marginalization of older states. For high-altitude operation, the configuration is tuned for reduced parallax and limited scene texture, tracking up to 150 features per frame with Kanade-Lucas-Tomasi (KLT) optical flow. Noise parameters are calibrated via Allan variance analysis, and the estimator initializes through joint visual-inertial structure-from-motion to recover metric scale.

\subsection{Visual Geo-Localization Module}

\subsubsection{Deep Feature Extraction and Matching}

The NGPS module extracts keypoints and 256-D descriptors per image using SuperPoint~\cite{superpoint}. Feature matching is performed using LightGlue~\cite{lightglue}, which employs adaptive inference depth (stopping early on easy pairs), point pruning (discarding unmatchable points), and bidirectional cross-attention for efficient and robust correspondence estimation. A minimum of $N_{min} = 20$ valid correspondences (above matchability threshold $\tau = 0.5$) is required for pose estimation.

\begin{figure}[t]
    \centering
    \includegraphics[width=\columnwidth]{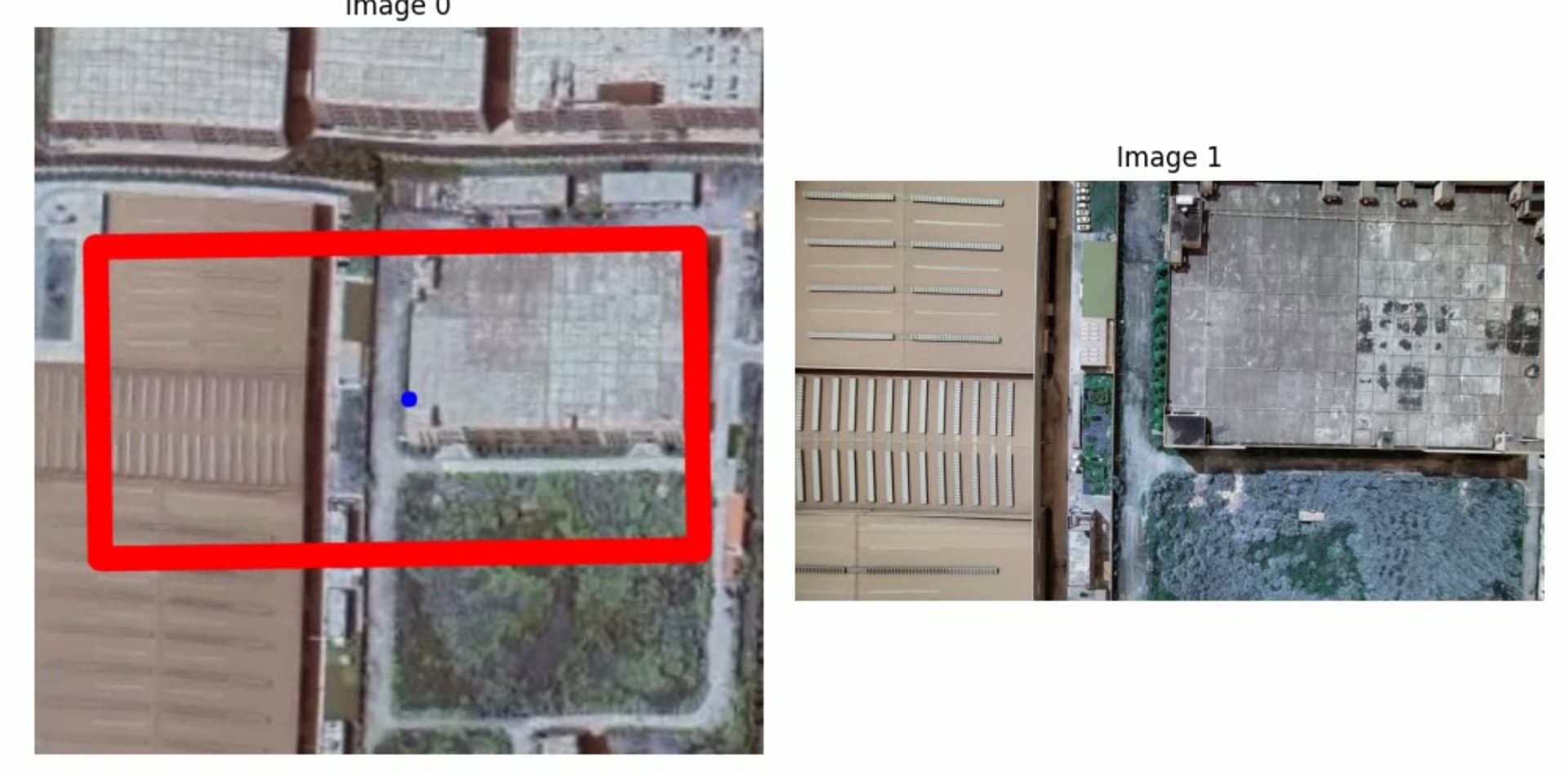}
    \caption{Deep feature matching: (left) satellite reference kernel with matched keypoints; (right) query image from the down-facing drone camera. SuperPoint keypoints and LightGlue correspondences enable pose estimation.}
    \label{fig:feature_matching}
\end{figure}

\subsubsection{Velocity-Predictive Kernel Extraction}
\label{sec:kernel}

To reduce computational load and search space, a kernel region of $k \times k$ pixels (default $k=300$) is extracted from the satellite reference image. Rather than centering on the last known position, which becomes stale at higher flight speeds given the 1--2\,Hz NGPS update rate - we predict the kernel center using the VIO velocity estimate from the UKF state:
\begin{equation}
\mathbf{p}_{kernel} = \mathbf{p}_{last} + \mathbf{S} \cdot \mathbf{R}^{wb} \hat{\mathbf{v}}^b \cdot \Delta t_{NGPS}
\label{eq:kernel_pred}
\end{equation}
where $\hat{\mathbf{v}}^b$ is the body-frame velocity from the UKF state vector, $\mathbf{R}^{wb}$ transforms it to the world frame, $\mathbf{S}$ converts meters to reference image pixels, and $\Delta t_{NGPS}$ is the time since the last NGPS update. This forward prediction ensures the kernel remains centered on the drone's current position even during fast traversals, significantly reducing matching failures that occur when the drone moves beyond the static kernel boundary between updates. The kernel is compensated for cumulative rotation $\theta_{tot}$:
\begin{equation}
\mathbf{R}_{kernel} = \begin{bmatrix} \cos\theta_{tot} & -\sin\theta_{tot} \\ \sin\theta_{tot} & \cos\theta_{tot} \end{bmatrix}
\end{equation}
and resized to $900 \times 900$ pixels. A homography $\mathbf{H}$ is estimated via RANSAC~\cite{ransac} (5-pixel reprojection threshold).

\subsection{Multi-Method Rotation Estimation}

Accurate rotation estimation is critical for position extraction and maintaining camera-to-reference alignment. We employ three complementary methods:

\textbf{Homography-Based:} The rotation angle is extracted from the normalized homography's upper-left $2 \times 2$ submatrix $\mathbf{R}_{norm}$ of $\mathbf{H}/H_{33}$:
\begin{equation}
\theta_{hom} = \text{atan2}(R_{norm,21},\; R_{norm,11})
\end{equation}
which approximates a 2D rotation under near-planar geometry.

\textbf{Keypoint-Based:} For each pair of matched keypoints $(i,j)$, vectors $\mathbf{v}_0^{(ij)}, \mathbf{v}_1^{(ij)}$ in reference and camera images yield rotation as the median angular difference:
\begin{equation}
\theta_{kpts} = \text{median}\!\left\{\text{atan2}\!\left(\mathbf{v}_0^{(ij)} \times \mathbf{v}_1^{(ij)},\; \mathbf{v}_0^{(ij)} \cdot \mathbf{v}_1^{(ij)}\right)\right\}
\end{equation}

\textbf{Contour-Based:} Camera image corners are transformed via $\mathbf{H}$ and a minimum-area rectangle is fitted, giving $\theta_{cnt}$ adjusted to $[-45^\circ, 45^\circ]$.

\textbf{Fusion and Validation:} The final estimate is the median of valid results: $\theta = \text{median}\{\theta_{hom}, \theta_{kpts}, \theta_{cnt}\}$. Temporal validation rejects estimates where $|\theta - \theta_{prev}| > 30^\circ$ or where the standard deviation over the last 3 samples exceeds $15^\circ$. Valid estimates are smoothed via weighted averaging over a 10-sample history buffer with linearly increasing weights.

\textbf{Position Extraction:} Position is estimated from the centroid of transformed camera corners, scaled from the kernel frame back to the reference frame:
\begin{equation}
\begin{bmatrix} x_{global} \\ y_{global} \end{bmatrix} = \begin{bmatrix} x_{base} \\ y_{base} \end{bmatrix} + \mathbf{R}^{-1}_{kernel} \begin{bmatrix} x_{offset} \\ y_{offset} \end{bmatrix}
\end{equation}

\subsection{UKF-Based Asynchronous Multi-Rate Sensor Fusion}

\subsubsection{State Definition}

The UKF maintains a 15-dimensional state vector:
\begin{equation}
\mathbf{x} = [x, y, z, \phi, \theta, \psi, \dot{x}, \dot{y}, \dot{z}, p, q, r, \ddot{x}, \ddot{y}, \ddot{z}]^T
\end{equation}
comprising position $(x,y,z)$, Euler angles $(\phi,\theta,\psi)$, body-frame velocity $(\dot{x},\dot{y},\dot{z})$, angular rates $(p,q,r)$, and body acceleration $(\ddot{x},\ddot{y},\ddot{z})$. Unlike a tightly-coupled IMU integrator, this UKF operates as a secondary kinematic fusion layer: IMU preintegration and bias estimation are handled entirely within VINS-Fusion~\cite{vins_fusion}, which supplies bias-compensated angular rates and accelerations as measurement inputs. The UKF state therefore tracks kinematic quantities rather than IMU error states, and no bias terms are needed. The system uses two IMUs: one onboard the flight controller for ArduPilot's EKF, and one offboard IMU used by both VIO (for preintegration) and the UKF (for direct attitude measurements).

\subsubsection{State Transition Model}

Position is updated by transforming body-frame velocity and acceleration to world frame: $\mathbf{p}_{k+1}^w = \mathbf{p}_k^w + \mathbf{R}_k^{wb} \mathbf{v}_k^b \Delta t + \tfrac{1}{2}\mathbf{R}_k^{wb} \mathbf{a}_k^b \Delta t^2$, with $\mathbf{R}_k^{wb}$ the body-to-world rotation. Velocity and Euler angles follow the standard kinematic update (constant acceleration, ZYX rate map).

\subsubsection{Process and Measurement Noise}

The process noise covariance $\mathbf{Q}$ is a diagonal matrix scaled by the prediction interval $\Delta t$, with per-state noise values tuned for high-altitude flight dynamics: position noise $\sigma^2_{x,y} = 10^{-6}$, altitude noise $\sigma^2_z = 10^{-7}$, orientation noise $\sigma^2_{\phi,\theta,\psi} = 10^{-8}$, velocity noise $\sigma^2_{\dot{x},\dot{y},\dot{z}} = 10^{-6}$, angular rate noise $\sigma^2_{p,q,r} = 10^{-8}$, and acceleration noise $\sigma^2_{\ddot{x},\ddot{y},\ddot{z}} = 10^{-4}$. The higher acceleration noise reflects the uncertainty in the constant-acceleration assumption during maneuvering.

\subsubsection{Measurement Models}

Three measurement models handle the multi-sensor inputs at different rates and dimensionalities:

\textbf{NGPS (2D):} Absolute position: $\mathbf{z}_{NGPS} = [x, y]^T$, providing the drift-bounding absolute reference at 1--2\,Hz with adaptive measurement noise (Section~\ref{sec:confidence}).

\textbf{VIO (6D):} Position and velocity: $\mathbf{z}_{VIO} = [x, y, z, \dot{x}, \dot{y}, \dot{z}]^T$, at 10--20\,Hz with covariance propagated from the VIO estimator's internal uncertainty.

\textbf{IMU (9D):} Orientation, angular velocity, acceleration: $\mathbf{z}_{IMU} = [\phi, \theta, \psi, p, q, r, \ddot{x}, \ddot{y}, \ddot{z}]^T$, at 100--200\,Hz with noise from Allan variance calibration.

\subsubsection{Adaptive Confidence-Weighted NGPS Noise}
\label{sec:confidence}

A critical challenge in fusing low-frequency absolute measurements with high-frequency relative estimates is that the quality of NGPS position measurements varies significantly depending on scene texture, illumination, and the degree of aerial-satellite appearance overlap. Treating all NGPS updates with fixed measurement noise leads to suboptimal fusion: low-confidence matches inject noise while high-confidence matches are underweighted.

We propose an adaptive measurement noise model that dynamically scales the NGPS covariance based on three quality metrics computed during feature matching. For each NGPS measurement, we extract:
\begin{itemize}
    \item $r_{inlier} = N_{inlier} / N_{total}$: RANSAC inlier ratio
    \item $\bar{e}_{reproj}$: mean reprojection error of inliers (pixels)
    \item $\bar{c}_{match}$: mean LightGlue confidence of accepted matches
\end{itemize}

These are fused into a scalar confidence score:
\begin{equation}
\gamma = \alpha_1 \cdot r_{inlier} + \alpha_2 \cdot \bar{c}_{match} - \alpha_3 \cdot \tanh\!\left(\bar{e}_{reproj} / e_0\right)
\label{eq:confidence}
\end{equation}
where $\alpha_1 = 0.4$, $\alpha_2 = 0.4$, $\alpha_3 = 0.2$, and $e_0 = 5.0$ is a normalization constant. The $\tanh$ saturates the reprojection error penalty, preventing outlier frames from dominating. Under this weighting, $\gamma$ has a theoretical upper bound of $0.8$ (for $r_{inlier}=\bar{c}_{match}=1$ and $\bar{e}_{reproj}\to0$), so we clip to $[\gamma_{min}, \gamma_{max}]$ with $\gamma_{min}=0.1$ and $\gamma_{max}=0.8$ to prevent division by zero and enforce bounded scaling.

The NGPS measurement covariance is then scaled inversely with confidence:
\begin{equation}
\mathbf{R}_{NGPS} = \frac{\sigma^2_{base}}{\gamma^2} \cdot \mathbf{I}_2
\label{eq:adaptive_noise}
\end{equation}
where $\sigma^2_{base} = 1.0$\,m$^2$ is the nominal baseline noise. High-confidence matches ($\gamma \to 0.8$) receive low covariance ($\approx 1.56\times \sigma^2_{base}$), strongly correcting the state, while low-confidence matches ($\gamma \to \gamma_{min}$) receive inflated covariance ($100\times$ larger), allowing the UKF to rely more heavily on VIO propagation. Through the Kalman gain, larger $\mathbf{R}_{NGPS}$ directly downweights the NGPS correction during the update step. This prevents erroneous NGPS matches from corrupting the state estimate while ensuring high-quality matches are fully leveraged.

\subsubsection{Temporal Priority Queue}

The core novelty of our fusion architecture is the asynchronous temporal priority queue that processes measurements in strict chronological order regardless of arrival time or source rate. Algorithm~\ref{alg:fusion} details this procedure. All incoming measurements are timestamped and enqueued. At each update cycle, the UKF processes all queued measurements up to the current time, performing a predict-update step for each. This ensures low-frequency absolute corrections (NGPS at 1--2\,Hz) are optimally interleaved with high-frequency relative estimates (VIO at 10--20\,Hz), maintaining temporal consistency.

\begin{algorithm}[t]
\caption{Confidence-Weighted Asynchronous Multi-Rate Fusion}
\label{alg:fusion}
\begin{algorithmic}[1]
\REQUIRE Streams: NGPS (${\sim}$1--2\,Hz), VIO (${\sim}$10--20\,Hz), IMU (${\sim}$100--200\,Hz)
\ENSURE Drift-corrected state $\hat{\mathbf{x}}$ at VIO rate
\STATE Initialize UKF: $\hat{\mathbf{x}}_0,\, \mathbf{P}_0$; priority queue $\mathcal{Q} \leftarrow \emptyset$
\WHILE{system active}
\STATE Enqueue new measurements into $\mathcal{Q}$ by timestamp
\FOR{each $z_i \in \mathcal{Q}$ with $t_i \leq t_{\text{now}}$}
\STATE \textbf{Predict:} $\hat{\mathbf{x}}^{-}\!,\, \mathbf{P}^{-} \leftarrow f(\hat{\mathbf{x}},\, \mathbf{P},\, t_i - t_{\text{last}})$
\IF{$z_i$ is NGPS}
\STATE Compute $\gamma$ from $r_{inlier},\, \bar{c}_{match},\, \bar{e}_{reproj}$ \hfill (Eq.~\ref{eq:confidence})
\STATE $\mathbf{R}_{NGPS} \leftarrow (\sigma^2_{base} / \gamma^2)\, \mathbf{I}_2$ \hfill (Eq.~\ref{eq:adaptive_noise})
\STATE \textbf{Update} with $h_{\text{NGPS}}$, noise $\mathbf{R}_{NGPS}$
\STATE $\mathbf{p}_{kernel} \leftarrow \mathbf{p}_{NGPS} + \mathbf{S}\,\mathbf{R}^{wb}\hat{\mathbf{v}}^b \Delta t$ \hfill (Eq.~\ref{eq:kernel_pred})
\ENDIF
\IF{$z_i$ is VIO}
\STATE \textbf{Update} with $h_{\text{VIO}}$, noise from VIO covariance
\ENDIF
\IF{$z_i$ is IMU}
\STATE \textbf{Update} with $h_{\text{IMU}}$, noise from Allan cal.
\ENDIF
\STATE $t_{\text{last}} \leftarrow t_i$
\ENDFOR
\STATE Publish $\hat{\mathbf{x}}$ to ArduPilot EKF
\ENDWHILE
\end{algorithmic}
\end{algorithm}

Angle wrapping uses circular mean and atan2-based differences so $\Delta\theta \in [-\pi,\pi]$.

\subsection{2.5D Georeferenced Orthomosaic Reconstruction}

The globally optimized camera poses enable real-time 2.5D georeferenced orthomosaic reconstruction as a byproduct of the localization pipeline. The VINS global optimization module performs pose graph optimization using Ceres Solver, incorporating both VIO relative pose constraints and NGPS absolute position anchors:
\begin{equation}
\min_{\{P_k\}} \sum_{k} \lVert P_k \ominus P_{k-1} - \Delta P_k^{\text{VIO}} \rVert^2_{\Sigma_{\text{VIO}}} + \sum_{k \in \mathcal{A}} \rho\!\left(\lVert \mathbf{t}_k - \mathbf{p}_k^{\text{NGPS}} \rVert^2_{\Sigma_{\text{NGPS}}}\right)
\end{equation}
where $\mathcal{A}$ is the set of keyframes with NGPS measurements and $\rho(\cdot)$ is the Huber loss for outlier robustness.

Given optimized pose $P_k = (\mathbf{R}_k, \mathbf{t}_k)$ and altitude $h_k$, each frame is projected onto the ground plane. The image-to-ground homography for frame $k$ is
\begin{equation}
\mathbf{H}_k = \mathbf{K} \begin{bmatrix} \mathbf{r}_1^k & \mathbf{r}_2^k & \mathbf{t}_k \end{bmatrix}
\end{equation}
where $\mathbf{r}_1^k, \mathbf{r}_2^k$ are the first two columns of $\mathbf{R}_k$ (world-to-camera), $\mathbf{t}_k$ is the translation, and $\mathbf{K}$ is the camera intrinsic matrix. The inverse $\mathbf{H}_k^{-1}$ maps image pixels to georeferenced ground coordinates. Frames are warped and blended with distance-to-center weights; elevation is averaged similarly to form the DSM. The UKF does not receive ArduPilot's z; only horizontal position and velocity are sent to the flight controller. For 2.5D, the companion may use ArduPilot's terrain-corrected altitude (baro, rangefinder, or SRTM) when available; otherwise $h_k$ from UKF state.

\begin{figure}[t]
    \centering
    \begin{tabular}{@{}c@{\hspace{0.5em}}c@{}}
    \includegraphics[width=0.48\columnwidth,height=4cm,keepaspectratio]{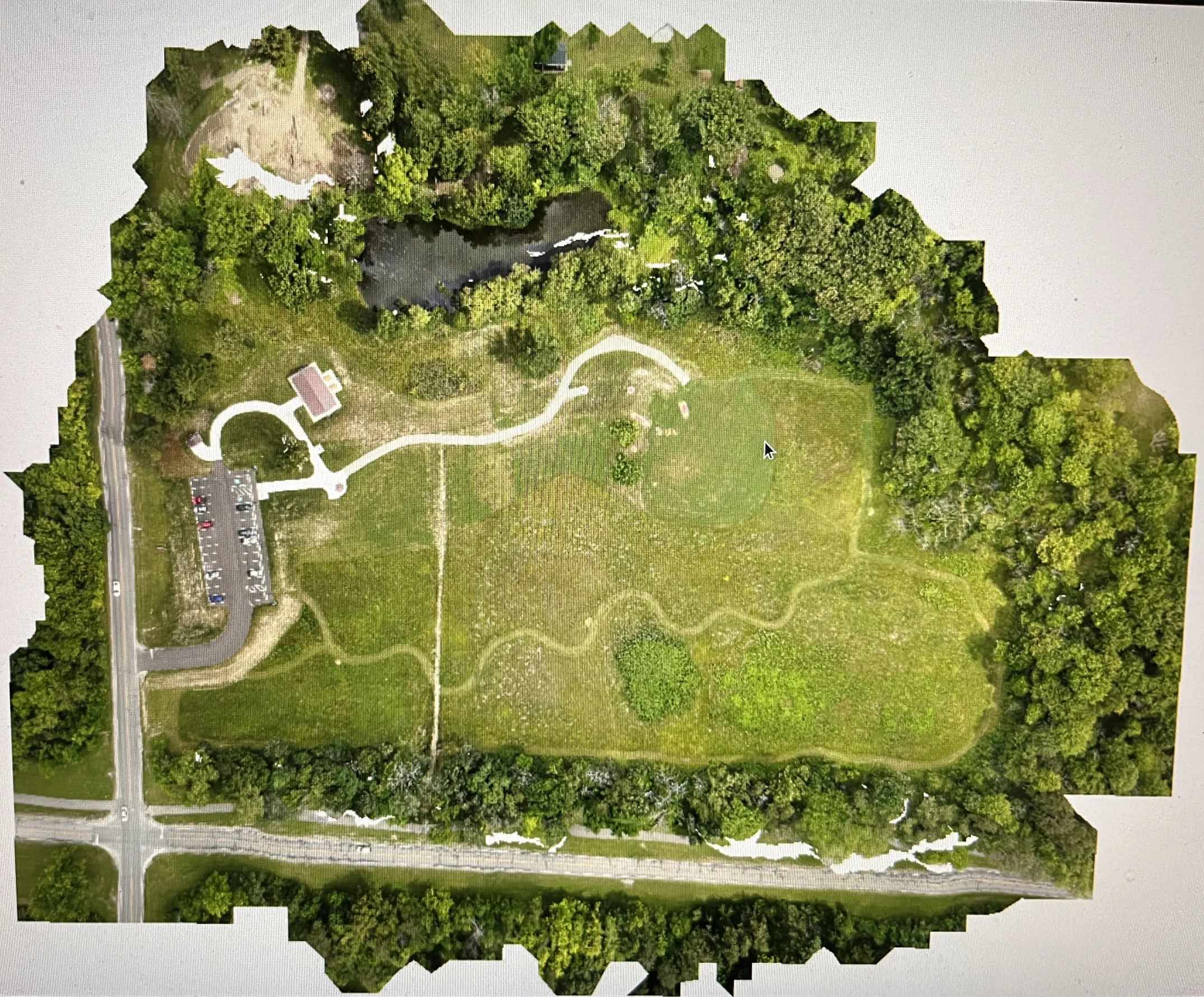} &
    \includegraphics[width=0.48\columnwidth,height=4cm,keepaspectratio]{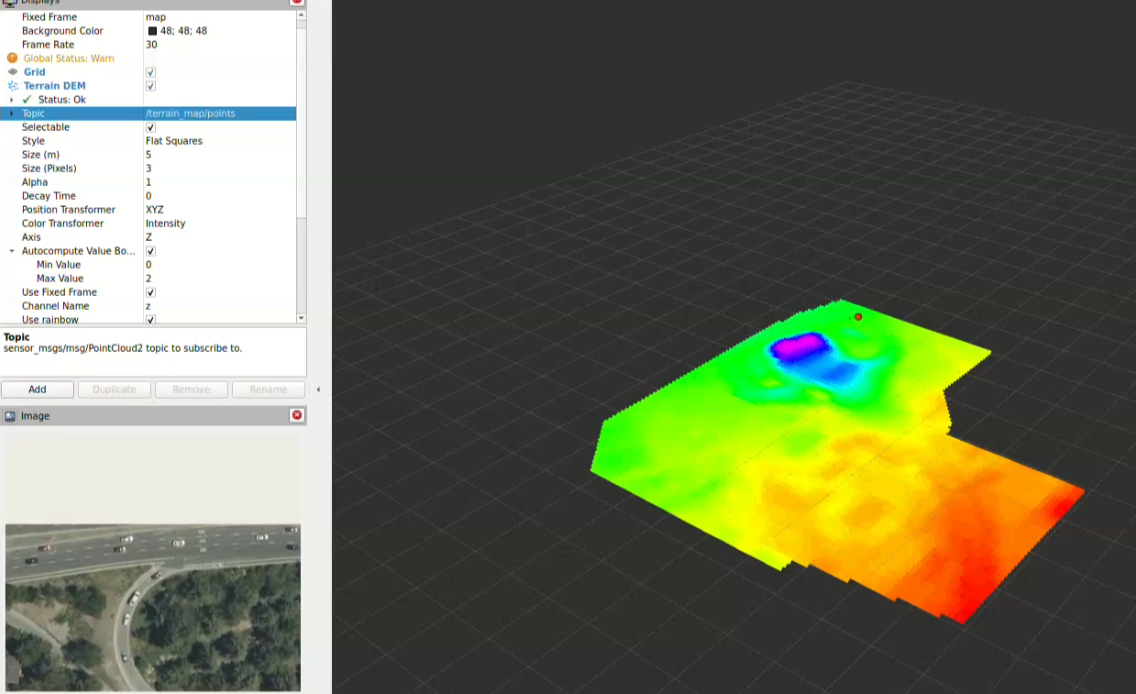} \\
    \small (a) Orthomosaic & \small (b) Real-time DSM
    \end{tabular}
    \caption{2.5D georeferenced reconstruction: (a) Orthomosaic post-processed from globally optimized poses (NGPS-anchored VINS), aligned to satellite reference; (b) Real-time digital surface model (DSM) and terrain visualization in RViz during flight.}
    \label{fig:orthomosaic}
\end{figure}

The global optimization module runs in a separate thread using Ceres Solver with Huber loss and sparse Cholesky factorization. VIO relative pose factors constrain consecutive poses ($\sigma_t = 0.1$\,m, $\sigma_q = 0.01$\,rad), while NGPS factors anchor absolute positions weighted by matching confidence. The VIO-to-global transform is updated after each optimization pass. The reconstruction operates incrementally: each new camera frame is projected and blended into the running mosaic - enabling real-time temporal change detection against the satellite reference and elevation-aware path planning via the digital surface model. Figure~\ref{fig:orthomosaic} shows an example orthomosaic post-processed from a flight dataset using the pipeline above.

\subsection{Multi-Source Altitude Estimation and Flight Control Integration}

The UKF outputs fused odometry at 10--20\,Hz. Only horizontal position $(x,y)$ and horizontal velocity are transmitted to ArduPilot via the uXRCE-DDS interface (\texttt{VISION\_POSITION\_ESTIMATE}); altitude remains estimated on the flight controller. ArduPilot's EKF3~\cite{ardupilot} operates at 100--400\,Hz, performing final sensor fusion for flight control.

A key aspect of the altitude estimation is the multi-source architecture within ArduPilot's EKF3, which provides accurate height-above-ground (HAG) across the full operating range (40--100+\,m). Three complementary altitude sources are fused:
\begin{itemize}
    \item \textbf{Barometric altimeter:} Provides the absolute altitude reference (MSL), always active, used for long-term altitude stability.
    \item \textbf{Rangefinder:} Provides precise ground-relative altitude when within sensor range (typically $<$40\,m), used as the primary HAG source at lower altitudes.
    \item \textbf{SRTM terrain database:} When the vehicle climbs beyond rangefinder range, the EKF3 automatically falls back to the Shuttle Radar Topography Mission (SRTM) database~\cite{srtm} for terrain-relative altitude estimation, enabling consistent HAG estimation up to 100+\,m without rangefinder coverage.
\end{itemize}

The automatic switching between rangefinder and terrain database is controlled by the EKF3's internal validity checks: if the rangefinder's ground offset estimate becomes invalid (beyond range), and SRTM terrain data has been received within 5\,s, the terrain altitude replaces the rangefinder as the HAG source. This is particularly critical for our operating range, where flights routinely exceed rangefinder limits. The companion sends only horizontal position and velocity to ArduPilot (no z); the UKF's altitude state is driven solely by VIO and IMU. When available, ArduPilot's terrain-corrected altitude can be used on the companion for the 2.5D ground-plane projection only, improving orthomosaic and DSM accuracy without affecting the UKF or VINS state.

\section{Experimental Evaluation}

\subsection{Experimental Platform and Setup}

\begin{figure}[t]
    \centering
    \includegraphics[width=0.75\columnwidth]{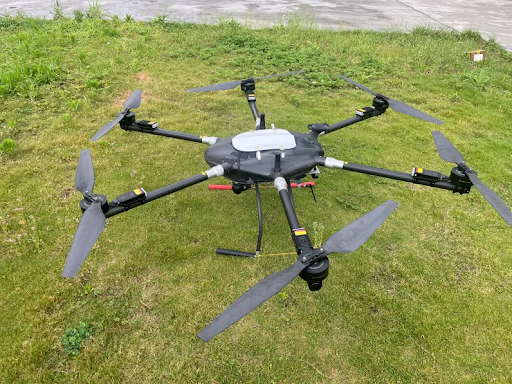}
    \caption{Experimental platform: custom hexacopter with Cube Orange Plus, Jetson Orin NX, and down-facing Topotek gimbal camera.}
    \label{fig:setup}
\end{figure}

The system was evaluated on a custom hexacopter platform with:
\begin{itemize}
    \item \textbf{Flight Controller:} Cube Orange Plus (Hex) with integrated IMU
    \item \textbf{Compute:} NVIDIA Jetson Orin NX (Ubuntu 22.04, CUDA 11.4)
    \item \textbf{Camera:} Topotek gimbal camera (down-facing, stabilized)
    \item \textbf{Software:} ROS2 Humble, PyTorch 2.0, ArduPilot with uXRCE-DDS; IMU-triggered camera frames via DDS provide hardware-level IMU--camera sync; fused state estimates sent back via the same interface
    \item \textbf{Reference:} Google Earth satellite imagery (${\sim}$0.3\,m/pixel GSD)
    \item \textbf{Ground Truth:} RTK-GPS ($\pm$2\,cm accuracy)
\end{itemize}

\begin{figure}[t]
    \centering
    \includegraphics[width=0.7\columnwidth]{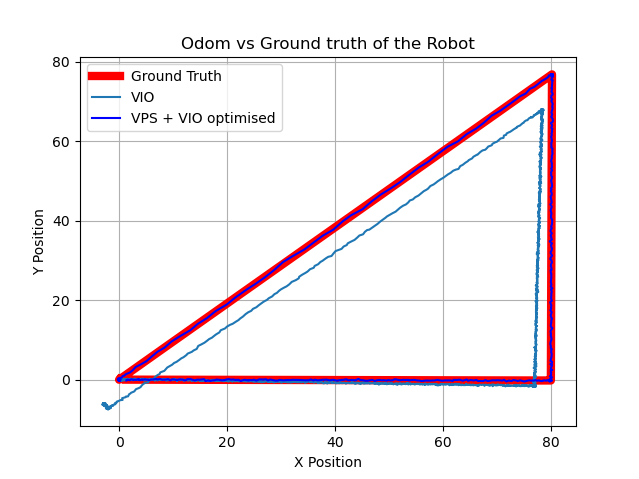}
    \caption{Gazebo+ArduPilot SITL validation: fused NGPS+VIO trajectory closely tracks ground truth while VIO-only drifts, confirming the multi-rate fusion architecture prior to hardware deployment.}
    \label{fig:sim_trajectory}
\end{figure}

Prior to real flights, the system was validated in Gazebo with ArduPilot SITL (Fig.~\ref{fig:sim_trajectory}), confirming the fused trajectory tracks ground truth while VIO-only drifts.

Calibration followed a three-step procedure. \textbf{Mono intrinsics:} Kalibr~\cite{kalibr} in Docker (ROS~1) with the camera stationary and a moving AprilGrid, yielding camera intrinsic parameters. \textbf{IMU noise:} Allan deviation analysis (allan\_ros2) on a 2-hour stationary rosbag to estimate white noise and bias instability for accelerometer and gyroscope~\cite{allan_variance}, yielding the IMU noise model. \textbf{IMU--camera extrinsics:} Kalibr in Docker with a rosbag in which the AprilGrid was fixed and the sensor module was moved to excite all axes, yielding extrinsics from IMU to camera frame. Five flight sequences were collected across diverse environments and altitudes (60--150\,m AGL):

\begin{table}[t]
\centering
\caption{Evaluation Dataset Characteristics}
\label{tab:datasets}
\footnotesize
\begin{tabular}{lcccc}
\toprule
\textbf{Sequence} & \textbf{Alt. (m)} & \textbf{Speed} & \textbf{Dur.} & \textbf{Environment} \\
\midrule
Urban-100  & 100 & 3\,m/s & 8.2\,min  & Urban rooftops \\
Suburb-80  & 80  & 3\,m/s & 10.5\,min & Suburban \\
Agri-60    & 60  & 4\,m/s & 6.1\,min  & Agricultural \\
Mixed-100  & 100 & 4\,m/s & 12.3\,min & Mixed terrain \\
High-150   & 150 & 2\,m/s & 9.8\,min  & Urban/mixed \\
\bottomrule
\end{tabular}
\end{table}

\subsection{Localization Accuracy}

\begin{figure}[t]
    \centering
    \includegraphics[width=0.62\columnwidth,height=6.5cm]{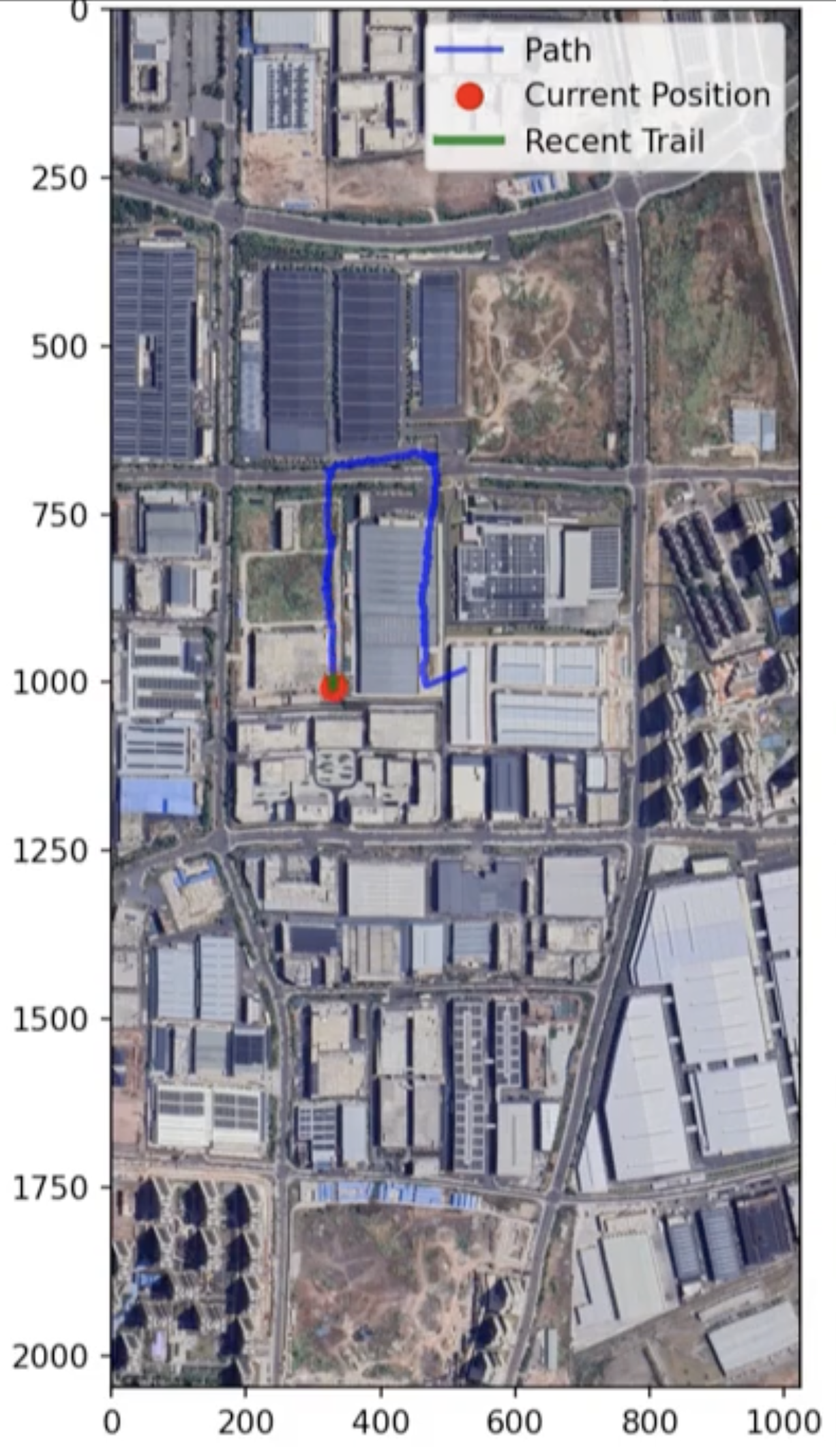}
    \caption{Fused NGPS+VIO trajectory overlaid on the satellite reference. The path is drift-corrected by periodic absolute position updates from visual geo-localization.}
    \label{fig:trajectory}
\end{figure}

Table~\ref{tab:position_results} summarizes position accuracy aggregated across all five sequences (60--150\,m AGL). The VIO-only baseline exhibits significant drift accumulation, particularly in the agricultural sequence where repetitive texture reduces feature discriminability. The NGPS-only system provides absolute position bounded by satellite image resolution but at low frequency with occasional matching failures (overall success rate: 87.0\%). The full system with adaptive confidence weighting achieves 2.94\,m RMSE, a $3.5\times$ improvement over VIO-only.

\begin{table}[t]
\centering
\caption{Position Error Statistics, meters (all five sequences)}
\label{tab:position_results}
\footnotesize
\begin{tabular}{lcccc}
\toprule
\textbf{Method} & \textbf{Mean} & \textbf{Std} & \textbf{Max} & \textbf{RMSE} \\
\midrule
VIO-only              & 9.12  & 4.89  & 31.4   & 10.37 \\
NGPS-only             & 3.91  & 2.14  & 9.84   & 4.46 \\
Fused (fixed noise)   & 2.94  & 1.48  & 7.82   & 3.29 \\
\textbf{Fused (adaptive, Ours)} & \textbf{2.61} & \textbf{1.32} & \textbf{6.87} & \textbf{2.94} \\
\bottomrule
\end{tabular}
\end{table}

Table~\ref{tab:per_sequence} provides the per-sequence breakdown with adaptive confidence weighting. The High-150 sequence dominates the aggregate because 150\,m AGL reduces satellite feature discriminability (larger ground footprint, lower effective resolution). The adaptive scheme provides the largest relative benefit in the agricultural sequence, where low-confidence matches from repetitive texture are automatically downweighted (RMSE 3.41\,m fixed $\to$ 2.73\,m adaptive). Urban scenes with high-quality matches benefit more modestly (RMSE 2.14\,m $\to$ 1.82\,m).

\begin{table}[t]
\centering
\caption{Per-Sequence Results (Adaptive Confidence Weighting)}
\label{tab:per_sequence}
\footnotesize
\begin{tabular}{lccccc}
\toprule
\textbf{Sequence} & \textbf{RMSE} & \textbf{Mean} & \textbf{Max} & \textbf{Success} & \textbf{VIO RMSE} \\
 & (m) & (m) & (m) & (\%) & (m) \\
\midrule
Urban-100  & 1.82 & 1.61 & 3.94 & 92.4 & 8.14 \\
Suburb-80  & 1.94 & 1.72 & 4.27 & 90.8 & 9.67 \\
Agri-60    & 2.73 & 2.41 & 5.88 & 83.6 & 15.32 \\
Mixed-100  & 2.06 & 1.83 & 4.51 & 89.7 & 8.89 \\
High-150   & 6.04 & 5.32 & 6.87 & 78.3 & 11.23 \\
\midrule
\textbf{Overall} & \textbf{2.94} & \textbf{2.61} & \textbf{6.87} & \textbf{87.0} & \textbf{10.37} \\
\bottomrule
\end{tabular}
\end{table}

When NGPS matching fails (21.7\% of frames in High-150, vs.\ 10.4\% on average for 60--100\,m sequences), the UKF continues with VIO propagation; the adaptive scheme ensures that the last few low-confidence matches before failure do not corrupt the state, providing a cleaner handoff to VIO-only mode.

The improvement factor (VIO RMSE / Fused RMSE) is $3.5\times$ overall, with the highest gain in Agri-60 ($5.6\times$) and the lowest in High-150 ($1.9\times$). The multi-method rotation estimation achieves robust orientation estimates, with the median fusion outperforming each individual method by rejecting single-method outliers (Table~\ref{tab:rotation_results}). The contour-based method exhibits the highest variance due to sensitivity to non-rectangular transformed footprints, while the homography-based method provides the most stable individual estimates.

\begin{table}[t]
\centering
\caption{Rotation Error Statistics (degrees)}
\label{tab:rotation_results}
\footnotesize
\begin{tabular}{lcccc}
\toprule
\textbf{Method} & \textbf{Mean} & \textbf{Std} & \textbf{Max} & \textbf{RMSE} \\
\midrule
Homography-based  & 2.14  & 1.87  & 12.3  & 2.84 \\
Keypoint-based    & 2.67  & 2.31  & 15.1  & 3.53 \\
Contour-based     & 3.42  & 2.98  & 18.7  & 4.54 \\
Fused (median)    & 1.58  & 1.12  & 6.8   & 1.94 \\
\bottomrule
\end{tabular}
\end{table}

\subsection{Comparison with Monocular VIO Baselines}

Table~\ref{tab:mono_comparison} compares our fused system against monocular VIO (VINS-Fusion, mono camera + IMU) results from George~et~al.~\cite{george2023vinsfusion}, who reported Absolute Trajectory Error (ATE) and Relative Error (RE) of yaw across multiple altitude and speed conditions. The mono-VIO at 40\,m / 2\,m/s diverged entirely. We additionally evaluate our system at 150\,m AGL (beyond the George~et~al. dataset range) - where ATE rises to 6.04\,m due to reduced satellite-image feature discriminability at larger footprints.

\begin{table}[t]
\centering
\caption{Comparison with mono-VIO~\cite{george2023vinsfusion}. ATE in meters, yaw RE in degrees. $^\dagger$Not in George et al.\ dataset; our evaluation only.}
\label{tab:mono_comparison}
\footnotesize
\begin{tabular}{cccccc}
\toprule
\textbf{v} & \textbf{h} & \multicolumn{2}{c}{\textbf{Mono-VIO}~\cite{george2023vinsfusion}} & \multicolumn{2}{c}{\textbf{Ours}} \\
\cmidrule(lr){3-4}\cmidrule(lr){5-6}
(m/s) & (m) & ATE & RE & ATE & RE \\
\midrule
2 & 40  & --     & --     & 2.41 & 1.62 \\
2 & 60  & 55.81  & 10.39  & 1.94 & 1.28 \\
2 & 80  & 12.09  & 1.63   & 1.74 & 1.18 \\
2 & 100 & 18.41  & 2.58   & 2.06 & 1.42 \\
2 & 150$^\dagger$ & --  & --  & 6.04 & 2.83 \\
\midrule
3 & 40  & 9.46   & 3.58   & 2.63 & 1.74 \\
3 & 60  & 16.55  & 2.18   & 2.18 & 1.47 \\
3 & 80  & 6.74   & 4.48   & 1.88 & 1.31 \\
3 & 100 & 20.30  & 4.01   & 2.31 & 1.58 \\
\midrule
4 & 40  & 9.96   & 2.32   & 2.91 & 1.93 \\
4 & 60  & 15.50  & 4.88   & 2.47 & 1.67 \\
4 & 80  & 20.26  & 2.27   & 2.12 & 1.44 \\
4 & 100 & 17.91  & 6.16   & 2.39 & 1.62 \\
\bottomrule
\end{tabular}
\end{table}

To enable direct comparison with the George et al. baseline~\cite{george2023vinsfusion}, we additionally evaluated our system at their reported test conditions, including 40\,m AGL, which falls below our primary operating range (60--150\,m).

We compare against the monocular VIO configuration from George et al.~\cite{george2023vinsfusion}, which matches our single down-facing camera constraint. While George et al. also tested stereo-VIO with superior performance, our platform uses monocular visual-inertial odometry, making their mono-VIO results the appropriate baseline for direct comparison under matched hardware constraints.

Our system bounds ATE within 1.7--6.0\,m (40--150\,m altitude), whereas mono-VIO varies 6.74--55.81\,m and fails at 40\,m/2\,m/s. The 6.04\,m worst case at 150\,m reflects reduced satellite feature discriminability; yaw RE stays bounded ($2.83^\circ$). ATE increases modestly at higher speeds; the velocity-predictive kernel (Section~\ref{sec:kernel}) and absolute NGPS corrections maintain robustness in all tested conditions.

\subsection{Computational Performance}

Table~\ref{tab:computational} reports timing on the Jetson Orin NX.

\begin{table}[t]
\centering
\caption{Computational Performance (Jetson Orin NX)}
\label{tab:computational}
\footnotesize
\begin{tabular}{lcc}
\toprule
\textbf{Component} & \textbf{GPU (ms)} & \textbf{CPU (ms)} \\
\midrule
SuperPoint extraction  & 47   & 183 \\
LightGlue matching     & 318  & 1847 \\
Homography + rotation  & 8    & 8 \\
Processing overhead & 13 & 14 \\
Total NGPS pipeline    & 386  & 2052 \\
\midrule
\textbf{Achievable rate}  & \textbf{${\sim}$2.6\,Hz} & \textbf{${\sim}$0.5\,Hz} \\
UKF fusion output      & \multicolumn{2}{c}{10--20\,Hz} \\
ArduPilot EKF          & \multicolumn{2}{c}{100--400\,Hz} \\
\bottomrule
\end{tabular}
\end{table}

GPU acceleration is essential: the NGPS module achieves ${\sim}$2.6\,Hz with CUDA versus ${\sim}$0.5\,Hz on CPU. LightGlue dominates the computation (82\% of GPU time), with its adaptive depth mechanism terminating after 3--4 transformer layers on easy pairs instead of the full 9, reducing matching time by up to 40\%. The UKF processes asynchronous updates at negligible cost ($<$1\,ms per cycle), ensuring 10--20\,Hz fused output regardless of NGPS rate. Explicit GPU memory management and garbage collection per frame prevent memory accumulation during extended flights.

\subsection{Ablation Studies}

\subsubsection{Matching Threshold Sensitivity}

Table~\ref{tab:ablation_threshold} shows matching threshold sensitivity; default 0.5 balances success rate and accuracy.

\begin{table}[t]
\centering
\caption{Effect of Matching Threshold on System Performance}
\label{tab:ablation_threshold}
\footnotesize
\begin{tabular}{cccc}
\toprule
\textbf{Threshold} & \textbf{Success (\%)} & \textbf{Avg. Matches} & \textbf{Fused RMSE (m)} \\
\midrule
0.3 & 96.8 & 312 & 3.74 \\
0.4 & 94.2 & 187 & 3.41 \\
\textbf{0.5} & \textbf{87.0} & \textbf{124} & \textbf{2.94} \\
0.6 & 78.3 & 68 & 3.28 \\
0.7 & 63.1 & 34 & 4.87 \\
\bottomrule
\end{tabular}
\end{table}

Lower thresholds (0.3) admit outliers; higher (0.7) cause frequent match failures and extended VIO-only propagation.

\subsubsection{Altitude-Dependent Performance}

RMSE increases with altitude: 2.73\,m at 60\,m, 1.82--2.06\,m at 80--100\,m, 6.04\,m at 150\,m. The 80--100\,m range gives near-optimal matching; 150\,m degrades due to larger ground footprint and reduced effective resolution.

\subsubsection{Adaptive Confidence Weighting}

Table~\ref{tab:ablation_confidence} compares NGPS noise strategies. Fixed noise treats all measurements equally; binary gating discards moderate-quality matches. Our adaptive weighting (Eq.~\ref{eq:adaptive_noise}) scales trust with match quality and achieves the best RMSE and max error.

\begin{table}[t]
\centering
\caption{Effect of NGPS Measurement Noise Strategy}
\label{tab:ablation_confidence}
\footnotesize
\begin{tabular}{lcc}
\toprule
\textbf{Strategy} & \textbf{Fused RMSE (m)} & \textbf{Max Error (m)} \\
\midrule
Fixed noise ($\sigma^2 = 1.0$)                & 3.29 & 7.82 \\
Binary gating ($\gamma > 0.5$)                & 3.11 & 8.14 \\
\textbf{Adaptive weighting (Ours)}            & \textbf{2.94} & \textbf{6.87} \\
\bottomrule
\end{tabular}
\end{table}

Adaptive weighting reduces RMSE by 11\% and max error by 12\% vs fixed noise.

\subsubsection{Component Contribution and Architecture}

Velocity-predictive kernel vs static centering: at 4\,m/s predictive maintains 84.7\% success vs 64.3\% for static (32\% improvement). Priority queue beats arrival-order (+20\%) and decimation (+35\%). Removing NGPS yields VIO-only (10.37\,m); removing VIO yields NGPS+IMU (4.46\,m).

\subsection{Limitations}

The system requires altitudes above ${\sim}$40\,m for sufficient aerial-to-satellite perspective similarity. Performance degrades in low-texture environments (e.g., agricultural fields) or when satellite imagery has significant temporal mismatch with current ground conditions. When NGPS matching fails, the UKF falls back to VIO-only propagation until successful re-lock. The kernel size limits recovery from large position jumps, and GPU acceleration is essential for real-time performance.

\section{Conclusion}

We presented NGPS, a GPS-denied geo-localization system for high-altitude UAVs that achieves 2.94\,m position RMSE across five sequences (60--150\,m AGL) through deep satellite image matching and adaptive confidence-weighted UKF sensor fusion. The worst-case ATE of 6.04\,m occurs at 150\,m AGL and 2\,m/s, where reduced satellite feature discriminability limits correction quality; performance improves to 1.82--2.06\,m at 80--100\,m AGL. The adaptive measurement noise (modulated by RANSAC inlier ratio, reprojection error, and match confidence) provides 11\% improvement over fixed-noise fusion. The velocity-predictive kernel extraction maintains robust matching at higher flight speeds (32\% success-rate improvement at 4\,m/s). The temporal priority queue ensures optimal integration of heterogeneous measurement rates. Globally optimized poses further enable 2.5D georeferenced orthomosaic reconstruction. The complete system provides a $3.5\times$ accuracy improvement over standalone VIO and runs in real time on an embedded Jetson Orin NX.

Future work: domain-adapted features, adaptive kernel sizing, multi-scale matching for lower altitudes, and 2.5D-based temporal change detection.

\section*{Acknowledgment}
The authors thank the ArduPilot development community and Japan Drones Co., Ltd. for providing open-source tools and ecosystem support. The authors are especially grateful to Randy Mackay for guidance and mentorship on development and system integration.

The authors acknowledge Chongqing Zhongyue Aviation for providing access to their hexacopter platform and support during flight testing and data collection. The authors further thank Cao Bing and Li Bin for the support and assistance during experiments and data collection.



\begin{thebibliography}{99}

\bibitem{superpoint} D. DeTone, T. Malisiewicz, and A. Rabinovich, ``SuperPoint: Self-supervised interest point detection and description,'' in \textit{Proc. IEEE/CVF Conf. Comput. Vis. Pattern Recognit. Workshops}, 2018, pp. 224--236.

\bibitem{lightglue} P. Lindenberger, P.-E. Sarlin, and M. Pollefeys, ``LightGlue: Local feature matching at light speed,'' in \textit{Proc. IEEE/CVF Int. Conf. Comput. Vis.}, 2023, pp. 17627--17638.


\bibitem{vins_mono} T. Qin, P. Li, and S. Shen, ``VINS-Mono: A robust and versatile monocular visual-inertial state estimator,'' \textit{IEEE Trans. Robot.}, vol. 34, no. 4, pp. 1004--1020, 2018.

\bibitem{orbslam3} C. Campos et al., ``ORB-SLAM3: An accurate open-source library for visual, visual-inertial, and multimap SLAM,'' \textit{IEEE Trans. Robot.}, vol. 37, no. 6, pp. 1874--1890, 2021.

\bibitem{kim_satellite} J. Kim, S. Lee, and I. S. Kweon, ``Robust feature matching for aerial image registration,'' in \textit{Proc. Asian Conf. Comput. Vis.}, 2012, pp. 1--12.

\bibitem{huang_crossview} S. Huang, M. Wang, and S. Mao, ``Cross-view geo-localization with layer-to-layer transformer,'' in \textit{Proc. Adv. Neural Inf. Process. Syst.}, 2021, vol. 34, pp. 29009--29020.

\bibitem{shi_crossview} Y. Shi, L. Liu, X. Yu, and H. Li, ``Spatial-aware feature aggregation for cross-view image based geo-localization,'' in \textit{Proc. Adv. Neural Inf. Process. Syst.}, 2019, vol. 32, pp. 10090--10100.

\bibitem{chen_satellite} Z. Chen, S. Lam, A. Jacobson, and M. Milford, ``Convolutional neural network-based place recognition,'' in \textit{Proc. Australas. Conf. Robot. Autom.}, 2014, pp. 1--8.

\bibitem{lin_satellite} T.-Y. Lin, Y. Cui, S. Belongie, and J. Hays, ``Learning deep representations for ground-to-aerial geolocalization,'' in \textit{Proc. IEEE Conf. Comput. Vis. Pattern Recognit.}, 2015, pp. 5007--5015.

\bibitem{msckf} A. I. Mourikis and S. I. Roumeliotis, ``A multi-state constraint Kalman filter for vision-aided inertial navigation,'' in \textit{Proc. IEEE Int. Conf. Robot. Autom.}, 2007, pp. 3565--3572.

\bibitem{okvis} S. Leutenegger, S. Lynen, M. Bosse, R. Siegwart, and P. Furgale, ``Keyframe-based visual-inertial odometry using nonlinear optimization,'' \textit{Int. J. Robot. Res.}, vol. 34, no. 3, pp. 314--334, 2015.

\bibitem{vins_fusion} T. Qin, S. Cao, J. Pan, and S. Shen, ``A general optimization-based framework for global pose estimation with multiple sensors,'' \textit{arXiv preprint arXiv:1901.03642}, 2019.

\bibitem{george2023vinsfusion} A. George, N. Koivum{\"a}ki, T. Hakala, J. Suomalainen, and E. Honkavaara, ``Visual-inertial odometry using high flying altitude drone datasets,'' \textit{Drones}, vol. 7, no. 1, 36, 2023.

\bibitem{ukf} S. J. Julier and J. K. Uhlmann, ``Unscented filtering and nonlinear estimation,'' \textit{Proc. IEEE}, vol. 92, no. 3, pp. 401--422, 2004.

\bibitem{ransac} M. A. Fischler and R. C. Bolles, ``Random sample consensus: A paradigm for model fitting with applications to image analysis and automated cartography,'' \textit{Commun. ACM}, vol. 24, no. 6, pp. 381--395, 1981.

\bibitem{kalibr} J. Rehder, M. Gasser, R. Siegwart, and P. Furgale, ``Kalibr: A toolbox for camera-IMU calibration,'' in \textit{Proc. Int. Symp. Exp. Robot.}, 2014, pp. 1--16.

\bibitem{allan_variance} D. W. Allan, ``Statistics of atomic frequency standards,'' \textit{Proc. IEEE}, vol. 54, no. 2, pp. 221--230, 1966.

\bibitem{nex_uav} F. Nex and F. Remondino, ``UAV for 3D mapping applications: a review,'' \textit{Appl. Geomat.}, vol. 6, no. 1, pp. 1--15, 2014.

\bibitem{ardupilot} ArduPilot Dev Team, ``ArduPilot: Open source autopilot software,'' 2024. [Online]. Available: \url{https://ardupilot.org}

\bibitem{srtm} T. G. Farr et al., ``The Shuttle Radar Topography Mission,'' \textit{Rev. Geophys.}, vol. 45, no. 2, RG2004, 2007.

\end{thebibliography}
\end{document}